\documentclass{article}
\usepackage{ijcai24}

\usepackage{times}
\usepackage{soul}
\usepackage{url}
\usepackage[hidelinks]{hyperref}
\usepackage[utf8]{inputenc}
\usepackage[small]{caption}
\usepackage{graphicx}
\usepackage{amsmath}
\usepackage{amsthm}
\usepackage{booktabs}
\usepackage{algorithm}
\usepackage{algorithmic}
\usepackage[switch]{lineno}

\usepackage{multicol}
\usepackage{multirow}
\usepackage{amssymb}

\usepackage{amsmath}

\title{Recall, Retrieve and Reason: Towards Better In-Context Relation Extraction}

\author{
    Anonymous submission
}

\author{
Guozheng Li$^1$
\and
Peng Wang$^{1,2}$\thanks{Corresponding author} \and
Wenjun Ke$^{1,2}$\and
Yikai Guo$^3$\and
\\Ke Ji$^1$\and
Ziyu Shang$^1$\and
Jiajun Liu$^1$\And
Zijie Xu$^1$\\
\affiliations
$^1$School of Computer Science and Engineering, Southeast University\\
$^2$Key Laboratory of New Generation Artificial Intelligence Technology and Its \\
Interdisciplinary Applications (Southeast University), Ministry of Education \\
$^3$Beijing Institute of Computer Technology and Application\\
\emails
\{gzli, pwang, kewenjun, jiajliu, keji, ziyus1999, zijiexu\}@seu.edu.cn
}

\begin{document}

\maketitle

\begin{abstract}
Relation extraction (RE) aims to identify relations between entities mentioned in texts. Although large language models (LLMs) have demonstrated impressive in-context learning (ICL) abilities in various tasks, they still suffer from poor performances compared to most supervised fine-tuned RE methods. Utilizing ICL for RE with LLMs encounters two challenges: (1) retrieving good demonstrations from training examples, and (2) enabling LLMs exhibit strong ICL abilities in RE. On the one hand, retrieving good demonstrations is a non-trivial process in RE, which easily results in low relevance regarding entities and relations. On the other hand, ICL with an LLM achieves poor performance in RE while RE is different from language modeling in nature or the LLM is not large enough. In this work, we propose a novel recall-retrieve-reason RE framework that synergizes LLMs with retrieval corpora (training examples) to enable relevant retrieving and reliable in-context reasoning. Specifically, we distill the consistently ontological knowledge from training datasets to let LLMs generate relevant entity pairs grounded by retrieval corpora as valid queries. These entity pairs are then used to retrieve relevant training examples from the retrieval corpora as demonstrations for LLMs to conduct better ICL via instruction tuning. Extensive experiments on different LLMs and RE datasets demonstrate that our method generates relevant and valid entity pairs and boosts ICL abilities of LLMs, achieving competitive or new state-of-the-art performance on sentence-level RE compared to previous supervised fine-tuning methods and ICL-based methods.
\end{abstract}

\section{Introduction}
The emergence of large language models (LLMs) such as GPT-3~\cite{brown2020language} represent a significant advancement in natural language processing (NLP). Instead of following a pre-training then fine-tuning
pipeline~\cite{radford2019language,devlin2019bert,liu2019roberta,raffel2020exploring}, which fine-tunes a pre-trained model on a task-specific dataset in a fully-supervised manner, LLMs employ a new paradigm known as in-context learning (ICL)~\cite{brown2020language} which formulates an NLP task under the paradigm of language generation and makes predictions by learning from a few demonstrations. LLMs with ICL demonstrate impressive performance comparable to traditional methods even with limited examples~\cite{brown2020language,zhao2021calibrate}.

Despite the generally promising ICL performances~\cite{wei2022chain,arora2023ask,shang2024ontofact}, current ICL in relation extraction (RE)~\cite{li2024unlocking} task suffers from relatively poor performance. RE is a pivotal task in NLP~\cite{li2022fastre,wang2023fmlre,wang2023pascore,ji2023hierarchical,liu2024towards}, necessitating a profound comprehension of natural language, which involves identifying a pre-defined relation between a given entity pair mentioned in the input sentence or marking it as \texttt{NA} if no relation is identified. Given a test example input, ICL for RE prompts the input of LLMs with a few demonstrations retrieved from the training data and the test input itself, then LLMs generate the corresponding relation. Recent studies~\cite{jimenez-gutierrez-etal-2022-thinking,ma2023large} have revealed a significant performance gap in LLMs when apply ICL to the RE task. The main obstacles that utilizing ICL for RE with LLMs are two-fold: (i) the low relevance regarding entity and relation in the retrieved demonstrations for ICL~\cite{wan-etal-2023-gpt}, and (ii) the failure of utilizing LLMs with moderate size (less than 10B) compared to 175B GPT-3 for ICL~\cite{li2023revisiting}.

\begin{figure}[t]
	\centering
	\includegraphics[width=0.48\textwidth]{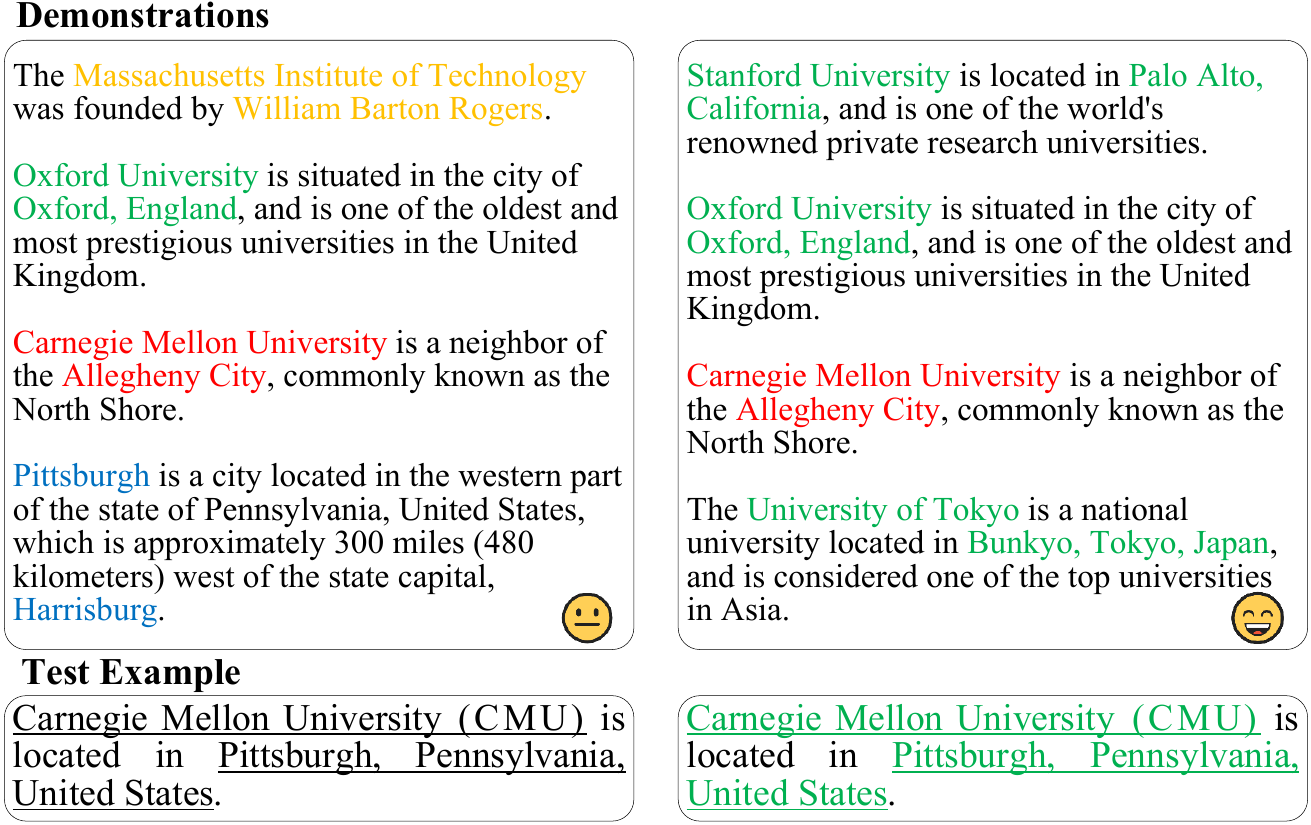}
	\caption{Comparison between naive demonstration selection (left) and our demonstration selection (right) methods. Different colors represent different relations between entity pairs, while green represents the golden relation expressed by entity pairs in test example.}
	\label{example}
\end{figure}

Typically, ICL demonstrations are selected randomly or via similarity-based sentence embedding obtained by sentence encoder such as Sentence-BERT~\cite{reimers2019sentence}. However, similarity-based retrieval is more concerned with the relevance of the overall sentence semantics and not as much with the specific entities and relations it contains, where consequently the test input tends to retrieve a semantically similar sentence but is not desired in terms of entities and relations. Similar with finding good demonstrations in ICL for general NLP tasks~\cite{liu-etal-2022-makes}, the demonstrations for RE should contain as much as similar or exactly same entities and relations regarding the test example, so as to significantly help the test example infer the relation between entities. However, current retrieval techniques~\cite{gao2021simcse,wan-etal-2023-gpt} fail to locate the delicate demonstrations considering the consistency of ontology (i.e., generalized entity types and the relations between them). Since entity types grouped under the similar ontology are often correlated and share similar relations~\cite{roche2003ontology}, we assume that examples sharing the similar ontology are relevant and could serve as good demonstrations for ICL. To this end, we propose to retrieve the relevant demonstrations in RE incorporating consistent ontological knowledge. Figure~\ref{example} illustrates the basic principles of our demonstration selection method, where we expect the demonstrations to include as many entity pairs as possible that express the same relation (i.e., similar ontology) with the test example. To obtain such ontological knowledge guided demonstrations, we propose to let the LLMs learn to generate relevant entity pairs that share the same relations with each test example, by implicitly acquiring the ontological knowledge of specific RE task during training phase. Then ICL demonstrations are retrieved from training examples by these recalled entity pairs. Compared to similarity-based retrieval, recalling entities explicitly via implicitly ontological knowledge guiding bring more benefits.

Besides the demonstration selection, the parameter scale of LLMs also plays an important role in ICL. Existing promising RE results obtaining via ICL all use very large-scale proprietary models~\cite{agrawal-etal-2022-large,li2023revisiting}. Although smaller open-source LLMs such as LLaMA~\cite{touvron2023llama} can also perform ICL similarly in simple classification tasks like sentiment classification, they are unable to perform more challenge tasks like extracting relations between entities through in-context demonstrations. Due to the limitations of proprietary models (higher computation cost and time consumption, concerns about privacy protection and local deployment, etc.), open-source smaller LLMs (less than 10B) achieve better ICL performance is of more applicability. Therefore, recent studies~\cite{chen2022meta,min2022metaicl} attempt to boost the moderate size LLMs ICL ability via meta in-context learning where an LLM is tuned to do in-context learning on a large set of training tasks. Inspired by this, we optimize LLMs to do in-context RE on training set, making the model becomes more effectively to reason about the relation between entities in-context by conditioning on a few training examples at inference time. Compared to directly optimize the input-output formats of examples (sentence and entity pair as input and relation label as output), this method can conduct in-context reasoning based on retrieved examples and generate more accurate results. 

To alleviate the above issues of irrelevant demonstration retrieval and inferior ICL capability, we present a recall-retrieve-reason framework, a novel RE method called \textbf{RE$^4$} (\textbf{R}elation \textbf{E}xtraction with \textbf{RE}call, \textbf{RE}trieve and \textbf{RE}ason) that synergizes open-source LLMs with retrieval corpora (training examples) to retrieve relevant demonstrations and conduct in-context reasoning. Specifically, RE$^4$ first generates entity pairs guided by ontological knowledge and grounded by retrieval corpora as valid queries via the recalling module. These entity pairs are then used to retrieve valid training examples from retrieval corpora in retrieval module to conduct in-context reasoning by reasoning module. In this way, we not only retrieve the relevant demonstrations from retrieval corpora but also consider the guidance of entity pairs and relations for reasoning. Based on this framework, RE$^4$ is joint optimized by two tasks: 1) recalling optimization, where we distill ontological knowledge from retrieval corpora into LLMs to generate relevant and valid entity pairs as queries; and 2) reasoning optimization, where we enable LLMs to conduct in-context reasoning based on retrieved demonstrations and generate predicted relations. We conduct extensive experiments on RE benchmarks to validate the effectiveness of RE$^4$. In summary, our contributions are three-fold: 
\begin{itemize}
    \item We propose a novel recall-retrieve-reason framework for RE that synergizes open-source LLMs with training examples to retrieve relevant demonstrations and conduct in-context reasoning. Moreover, the recalling module of RE$^4$ can be plug-and-play with different LLMs during inference to improve their performance.
    \item We distill the consistently ontological knowledge from training examples to guide LLMs for valid and relevant entity generation, and propose to boost the open-source LLMs in-context reasoning for RE via ICL tuning.
    \item We conduct extensive experiments on RE benchmarks and the results demonstrate that RE$^4$ achieves state-of-the-art performance in sentence-level RE.
\end{itemize}


\begin{figure*}[t]
	\centering
	\includegraphics[width=0.99\textwidth]{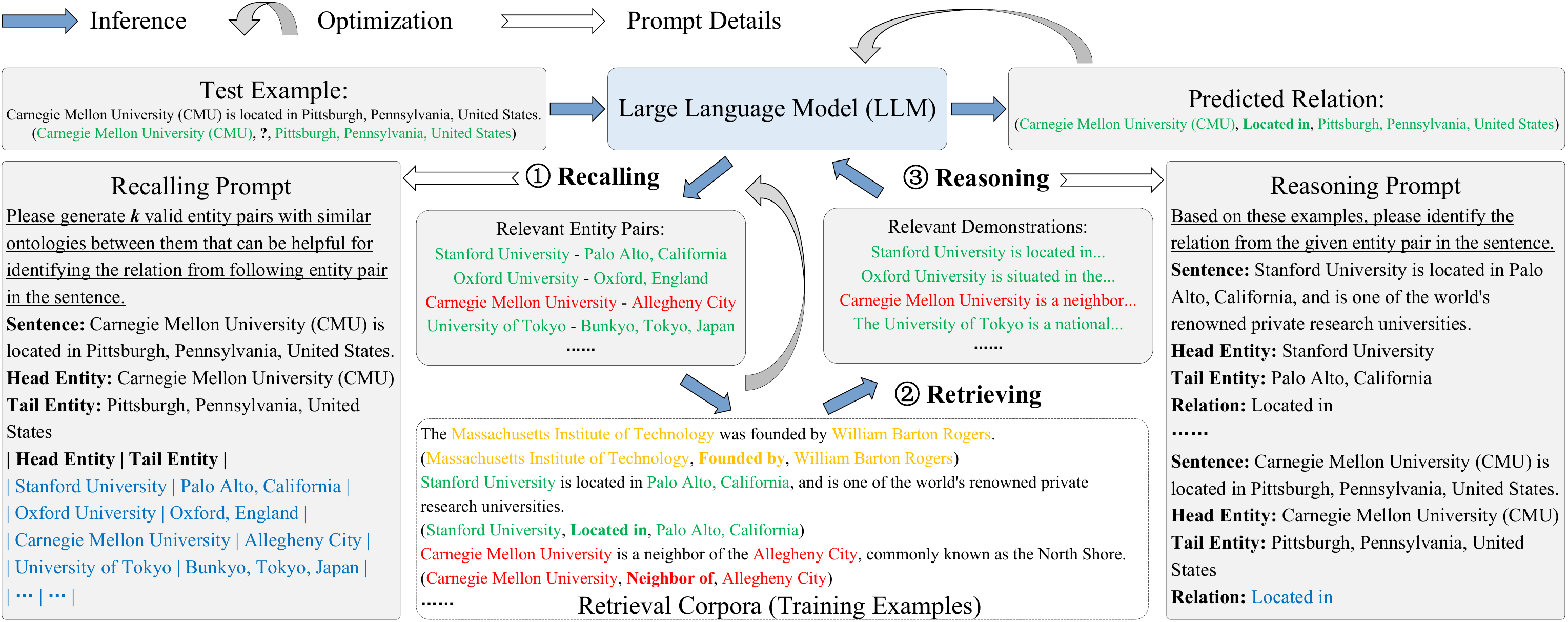}
	\caption{Illustration of the RE$^4$ framework. Given a test example, we first prompt LLMs to generate several relevant entity pairs that are grounded by retrieval corpora as queries. Then we retrieve demonstrations from training examples using the queries. Finally, we conduct in-context reasoning based on the retrieved entities and relations. The instructions in prompts are marked with \underline{underline}, and the outputs of LLMs in prompts are highlighted in blue.}
	\label{framework}
\end{figure*}

\section{Methodology}
\subsection{Task Formulations} 
For a given natural language sentence $s$ contains $N$ words, relation extraction aims at extracting the pre-defined relationship $r$ between $h$ and $t$, where $h$ and $t$ are two given target entities in $s$. If there is no pre-defined relation between $h$ and $t$, predict \texttt{NA}. Following previous works~\cite{cabot2021rebel,paolini2021structured,li2023sequence}, we treat RE as a generation task following the standard decoding manner. The optimization procedure can be written as follows:
\begin{equation}
    p(r|s,h,t) = \prod_{l=1}^L p(y_l|s,h,t,y_{<l})
\end{equation}
where we use an auto-regressive transformer~\cite{vaswani2017attention} decoder to
generate the relation name $r$ with label length $L$, and $p(y_l|s,h,t,y_{<l})$ denotes the probability of each token $y$ in $r$ generated by LLMs.

\subsection{Framework Overview}
Recently, many techniques have been explored to improve the ICL ability of LLMs in RE, which first retrieves relevant demonstrations from training examples and then conduct in-context RE based on them~\cite{wan-etal-2023-gpt}. However, the challenges of low-quality demonstration retrieval and inferior ICL ability of open-source models hinder the progress of ICL for RE in the era of LLMs. To address these issues, we propose a novel recall-retrieve-reason framework, which recalls relevant entity pairs, retrieves good demonstrations and then perform in-context reasoning for better relation predictions. The overall framework of RE$^4$ is illustrated in Figure~\ref{framework}. Given a test example ``\textit{Carnegie Mellon University (CMU) is located in Pittsburgh, Pennsylvania, United States}'', aiming to identify the relation between ``\textit{Carnegie Mellon University (CMU)}'' and ``\textit{Pittsburgh, Pennsylvania, United States}'', we generate an entity pair ``\textit{Stanford University}'' and ``\textit{Palo Alto, California}'' as the query. This entity pair expresses the similar relation with the test entity pair as the consistency of ontology between them (i.e., the head entity type is \textit{Organization} and the tail entity type is \textit{Location}). Then we retrieve the corresponding demonstration from training examples. Finally, we predict the relation (i.e. \textit{Located in}) based on demonstrations. 

\subsection{Framework Optimization}
We formulate our RE$^4$ as an optimization problem that aims to maximize the probability of reasoning the relation $r$ from a retrieval corpora $\mathcal{C}$ w.r.t the test example $e$ by generating consistently ontological entity pairs $z$ as the queries:
\begin{equation}
\label{eq2}
    p_{\theta}(r|e,\mathcal{C}) = \sum_{z \in \mathcal{Z}} p_{\theta}(r|e,z,\mathcal{C}) p_{\theta}(z|e)
\end{equation}
where $\theta$ denotes the parameters of LLMs, $z$ denotes the entity pairs (queries) generated by LLMs, and $\mathcal{Z}$ denotes the set of entity pairs that share similar ontology with the test example $e$. The latter term $p_{\theta}(z|e)$ is the probability of generating a valid entity pair $z$ grounded by retrieval corpora given $e$, which is realized by the recalling module. The former term $p_{\theta}(r|e,z,\mathcal{C})$ is the probability of reasoning the relation $r$ in context given the test example $e$, entity pair $z$, and retrieval corpora $\mathcal{C}$, computing by the reasoning module.

Despite the advantage of generating entity pairs as queries, the LLMs have zero ontological knowledge of the entities and relations contained in training examples. Therefore, LLMs cannot directly generate entity pairs grounded by retrieval corpora as valid queries. Moreover, LLMs might not have strong ICL ability to conduct effective in-context reasoning based on them. To address these issues, we design two instruction tuning tasks: 1) recalling optimization, which distills the consistent ontological knowledge from training examples into LLMs to generate valid entity pairs as queries, and 2) reasoning optimization, which enables LLMs to perform in-context reasoning based on retrieved demonstrations. The objective function in equation~(\ref{eq2}) is optimized by maximizing the evidence lower bound, which is formulated as:
\begin{equation}
    \begin{aligned}
        \log p_{\theta}(r|e,\mathcal{C}) \geq & \, \mathbb{E}_{z \sim q(z)} [\log p_{\theta}(r|e,z,\mathcal{C})] \\& -
        D_{\rm{KL}} (q(z) \, || \, p_{\theta}(z|e))
    \end{aligned}
\end{equation}
where $q(z)$ denotes the posterior distribution of valid and relevant entity pairs grounded by retrieval corpora. The latter term minimizes the KL divergence between the posterior and the prior, which encourages LLMs to generate consistently ontological entity pairs (i.e. recalling optimization). The former term maximizes the expectation that reasoning module generates correct relations based on the retrieved demonstrations (i.e. reasoning optimization).

\subsection{Recalling Entity Pairs}
To make LLMs generate valid entity pairs as queries for retrieving consistently ontological demonstrations from retrieval corpora, we minimize the KL divergence with the posterior distribution of valid entity pairs $q(z)$, which can be approximated by the valid entity pairs in retrieval corpora $\mathcal{C}$. 

Given a test example $e$ and its golden relation $r$, we could find the entity pair instances $z=(h, t)$ expressing the same relation $r$ in training examples. And $z$ can be considered valid and serve as a query for retrieving the relevant demonstration of $e$. And $q(z)$ can be formally approximated as:
\begin{equation}
    q(z) \cong q(z|r,e,\mathcal{C}) = \frac{1}{|\mathcal{Z}|}, \exists \, z \in \mathcal{C}
\end{equation}
where we assume a uniform distribution over all consistently ontological entity pairs $\mathcal{Z}$ regarding $e$, and $\exists \, z \in \mathcal{C}$ denotes the existence of an entity pair instance connecting $e$ and $r$ in $\mathcal{C}$. Therefore, the KL divergence can be calculated as:
\begin{equation}
    \mathcal{L}_{\rm {recall}} = D_{\rm {KL}} (q(z) \, || \, p_{\theta}(z|e)) \cong - \frac{1}{|\mathcal{Z}^{*}|} \sum_{z \in \mathcal{Z}^{*}} \log p_{\theta}(z|e)
\end{equation}
where we use the partial entity pairs $\mathcal{Z}^{*} \subset \mathcal{Z}$ relevant to $e$ in retrieval corpora $\mathcal{C}$ as supervision signals~\cite{luo2023reasoning}, where we distill the consistently ontological knowledge from training examples to LLMs. To utilize the instruction-following ability of LLMs~\cite{wei2022finetuned}, we design an instruction template that prompts LLMs to generate $k = |\mathcal{Z}^{*}|$ entity pairs. Therefore, the optimization of $\mathcal{L}_{\rm {recall}}$ becomes:
\begin{equation}
    - \frac{1}{|\mathcal{Z}^{*}|} \sum_{z \in \mathcal{Z}^{*}} \log p_{\theta}(z|e) = - \frac{1}{|\mathcal{Z}^{*}|} \sum_{z \in \mathcal{Z}^{*}} \log \prod_{i=1}^{|z|} p_{\theta} (y_i | y_{<i}, e)
\end{equation}
where $p_{\theta} (z|e)$ denotes the prior distribution of generating valid entity pair $z$, and $p_{\theta} (y_i | y_{<i}, e)$ denotes the probability of each token in $z$ generated by LLMs.

\subsection{Retrieving From Corpora}
Given a test example $e$ and an entity pair as query $z$, the retrieving module aims to retrieve the relevant demonstration $d$ from retrieval corpora. The retrieval process can be simply conducted by exact match between the generated entity pair $z$ and the entity pair of each candidate demonstration $d$:
\begin{equation}
    \mathcal{D} = \{d \, | \, h_{z} = h_{d}, t_{z} = t_{d}, z = (h_{z}, t_{z}), d = (s, h_{d}, t_{d}) \}
\end{equation}
where $\mathcal{D}$ denotes the set of retrieved demonstrations for ICL. Despite we can utilize the retrieved entity pairs to directly get the predicted relations via majority vote, the retrieved entity pairs of relations could be noisy and imperfect to the test example $e$, leading to incorrect predictions. Therefore, we use a reasoning module to boost the ICL ability of LLMs to identity the important entity pairs of relations and predict relations based on them via in-context reasoning~\cite{chen2023reckoning}.

\subsection{Reasoning Through Demonstrations}
In reasoning module, we aim to enable LLMs to conduct in-context reasoning based on the relevant entity pairs. And reasoning on multiple relevant entity pairs is formulated as:
\begin{equation}
    p_{\theta} (r | e, \mathcal{Z}, \mathcal{C}) = \prod_{z \in \mathcal{Z}} p_{\theta} (r | e, z, \mathcal{C})
\end{equation}
By approximating the expectation with $k$ sampled entity pairs $\mathcal{Z}^{*}$, the objective function of reasoning optimization where maximizes the probability of LLMs generating golden relations based on the relevant entity pairs is formalized as :
\begin{equation}
    \begin{aligned}
        \mathcal{L}_{\rm {reason}} &= - \mathbb{E}_{z \sim q(z)} [\log p_{\theta}(r|e,z,\mathcal{C})] \\& = - \sum_{z \in \mathcal{Z}^{*}} \log p_{\theta} (r|e,z,\mathcal{C}) \\& = - \log p_{\theta} (r|e,\mathcal{Z}^{*},\mathcal{C})
    \end{aligned}
\end{equation}

The reasoning module takes the test example $e$ and a set of retrieved demonstrations $\mathcal{D}$ to generate relation $r$. Similar with recalling module, we design a reasoning instruction prompt to guide LLMs to conduct in-context reasoning based on the retrieved demonstrations $\mathcal{D}$. The $\mathcal{D}$ are formulated as a series of structural sentences in standard ICL paradigm. The optimization of $\mathcal{L}_{\rm {reason}}$ is formulated as:
\begin{equation}
    \log p_{\theta} (r|e,\mathcal{Z}^{*},\mathcal{C}) = \log \sum_{z \in \mathcal{Z}^{*}} \sum_{d \in \mathcal{D}} \prod_{i=1}^{|r|} p_{\theta} (y_i | y_{<i},e,d)
\end{equation}
where $p_{\theta} (r|e,\mathcal{Z}^{*},\mathcal{C})$ denotes probability of reasoning the golden relation $r$ based on $k$ retrieved demonstrations $\mathcal{Z}^{*}$, and $y_i$ denotes the $i$-th token of relation $r$. To reduce the impact of error propagation during recalling process and improve the robustness of the model for in-context reasoning during inference time, we add some noise into $\mathcal{Z}^{*}$ by replacing $k^{*}, (1 \leq k^{*} \leq k)$ of $k$ demonstrations in $\mathcal{Z}^{*}$ with training examples share different relations from retrieval corpora with a uniform distribution. In this way, the LLMs can learn to in-context reasoning~\cite{chen2022meta,min2022metaicl,coda2023meta} based on important entities and relations, avoiding simply deduce the relations via majority vote.

\subsection{Joint Optimization and Inference}
The final objective function of RE$^4$ is the combination of the recalling optimization and reasoning optimization, which can be formulated as:
\begin{equation}
    \begin{aligned}
            \mathcal{L} &= - \frac{1}{|\mathcal{Z}^{*}|} \sum_{z \in \mathcal{Z}^{*}} \log \prod_{i=1}^{|z|} p_{\theta} (y_i | y_{<i}, e) \\&- \log \sum_{z \in \mathcal{Z}^{*}} \sum_{d \in \mathcal{D}} \prod_{i=1}^{|r|} p_{\theta} (y_i | y_{<i},e,d)
    \end{aligned}
\end{equation}
We adopt the same LLM for both recalling and reasoning, which are jointly trained on two instruction tuning tasks. To enhance the efficiency of the fine-tuning process and reduce memory requirements, we utilize Low-Rank Adaptation (LoRA)~\cite{hu2021lora}, which freezes the pre-trained model weights and injects trainable rank decomposition matrices into each layer of the Transformer~\cite{vaswani2017attention} architecture, greatly reducing the number of trainable parameters for downstream tasks. During inference, based on the recalled entity pairs and  retrieved demonstrations, the LLMs conduct ICL to generate predicted relations.

\section{Experiments}
\subsection{Settings}
\paragraph{Datasets and Metrics.} We evaluate RE$^4$ on SemEval 2010~\cite{hendrickx2019semeval}, TACRED~\cite{zhang2017position}, Google RE~\footnote{https://github.com/google-research-datasets/relation-extraction-corpus}, SciERC~\cite{luan2018multi}, four commonly used RE datasets. Following previous works~\cite{yamada2020luke,cabot2021rebel,li2023reviewing}, we use the Micro-F1 score excluding \texttt{NA} as the metric for evaluation. The statistics of datasets are shown in Table~\ref{data}. 

\begin{table}[h]
\small
\centering\setlength{\tabcolsep}{3mm}
\begin{tabular}{lcccc}
\toprule
{Dataset} & {\#Relation} & {\#Train} & {\#Dev} & {\#Test} \\
\midrule
{SemEval} & {9} & {6,507} & {1,493} & {2,717} \\
{TACRED} & {41} & {68,124} & {22,631} & {15,509} \\
{Google RE} & {5} & {38,112} & {9,648} & {9,616} \\
{SciERC} & {7} & {3,219} & {455} & {974} \\
\bottomrule
\end{tabular}
\caption{Statistics of datasets.}
\label{data}
\end{table}

\paragraph{Baselines.} We compare RE$^4$ with state-of-the-art RE models that represent a diverse array of approaches. Supervised fine-tuning (SFT) RE methods can be divided into three categories. Classification-based methods fine-tune language models on RE datasets with classification losses, such as MTB~\cite{baldini-soares-etal-2019-matching}, LUKE~\cite{yamada2020luke}, IRE~\cite{zhou-chen-2022-improved} and KLG~\cite{li2023reviewing}. Prompt-based methods use prompt and treats RE as a cloze-style task, such as KnowPrompt~\cite{chen2022knowprompt} and NLI-DeBERTa~\cite{sainz2021label}. Generative-based methods use text generation models for RE, such as REBEL~\cite{cabot2021rebel}, TANL~\cite{paolini2021structured}, RELA~\cite{li2023sequence} and DeepStruct~\cite{wang2022deepstruct}. For ICL-based RE, existing methods typically rely on the strong ICL ability of large-scale proprietary models without any SFT. We utilize GPT-RE~\cite{wan-etal-2023-gpt} which use PURE~\cite{zhong-chen-2021-frustratingly} as the demonstration retriever and GPT-3~\cite{brown2020language} as the base LLM for ICL. RE$^4$ can be regarded as the combination of SFT and ICL paradigms. On the one hand, it makes LLMs better adapt to specific tasks via instruction tuning. On the other hand, it improves the in-context reasoning ability of LLMs and enables relevant retrieved demonstrations during the ICL process. 

\paragraph{Experiment Details.} We experiment RE$^4$ with open-source LLMs including T5~\cite{raffel2020exploring}, BART~\cite{lewis-etal-2020-bart} and LLaMA~\cite{touvron2023llama}. For model scales, we select T5-Base (220M), T5-Large (770M), BART-Base (140M), BART-Large (400M) and LLaMA-7B for experiments. We utilize LoRA~\cite{hu2021lora} to tune LLMs for simplicity and efficiency. We set the rank $r$ of the LoRA parameters to 8 and the merging ratio $\alpha$ to 32. We train RE$^4$ for 5 epochs with batch size 4 and learning rate 1e-4. For the number of generated entity pairs, we set $k$ to 5. The checkpoint of LoRA adapter that achieves the best result on the validation set is used for testing. We also directly treat relation names as generation objectives and fine-tune the LLaMA.

\begin{table}[t]
\small
\centering\setlength{\tabcolsep}{0.7mm}
\begin{tabular}{c|c|c|c|c}
\toprule
\textbf{Method (\#Param.)} & \textbf{SemEval} & \textbf{TACRED} & \textbf{Google RE} & \textbf{SciERC} \\
\midrule
{MTB (336M)}  & {89.5} & {71.5} & {92.7} & {87.4} \\
{LUKE (355M)} & {90.1} & {72.7} & \underline{94.0} & {87.7} \\
{IRE (355M)} & {89.8} & {74.6} & {93.1} & {88.9} \\
{KLG (355M)} & {90.5} & {75.6} & {-} & {-} \\
\midrule
{KnowPrompt (355M)} & {90.2} & {72.4} & {-} & {-} \\
{NLI-DeBERTa (1.5B)} & {-} & {73.9} & {-} & {-} \\
\midrule
{REBEL (400M)} & {-} & {73.7} & {93.5} & {86.3} \\
{TANL (220M)} & {-} & {74.8} & {-} & {-} \\
{RELA (400M)} & {90.4} & {71.2} & {93.9} & \underline{90.3} \\
{DeepStruct (10B)} & {-} & \underline{76.8} & {-} & {-} \\
\midrule
{GPT-3 (175B)} & {70.1} & {32.5} & {-} & {-} \\
{GPT-RE (175B)} & \underline{91.9} & {72.1} & {-} & {-} \\
\midrule
{BART-Base w/ RE$^4$} & {89.8} & {71.5} & {92.4} & {86.0} \\
{BART-Large w/ RE$^4$} & {90.6} & {73.3} & {93.1} & {87.2} \\
{T5-Base w/ RE$^4$} & {89.9} & {72.7} & {92.6} & {86.3} \\
{T5-Large w/ RE$^4$} & {90.9} & {75.6} & {93.4} & {87.8} \\
{LLaMA w/ RE$^4$} & \textbf{92.1} & \textbf{77.2} & \textbf{94.5} & \textbf{91.7} \\
{LLaMA w/o RE$^4$} & {90.6} & {75.0} & {92.9} & {89.5} \\
\bottomrule
\end{tabular}
\caption{Micro-F1 score of test sets on four RE datasets. Results of baselines are retrieved from original papers. For ICL-based results, we use best 30-shot on SemEval and 15-shot on TACRED. Previous state-of-the-art results are marked with \underline{underline}, and best results are \textbf{bold}. Results of RE$^4$ are averaged over three random seeds.}
\label{re}
\end{table}

\subsection{Main Results}
The main results of baselines and RE$^4$ are summarized in Table~\ref{re}, where RE$^4$ with LLaMA outperforms all previous state-of-the-art methods on four RE datasets.
Compared to the classification-based methods, RE$^4$ shows favorable results without any external dataset usage or additional pre-training stages, while MTB and LUKE all involve entity and relation related pre-training tasks. RE$^4$ with most LLMs also suppresses two prompt-based methods. Compared to previous generative-based methods, RE$^4$ consistently achieves competitive or superior results on four datasets. For example, with same backbone BART-Large, RE$^4$ and REBEL share similar overall results, while REBEL is pre-trained with a large external relational triple extraction dataset. Notably, RE$^4$ with LLaMA significantly outperforms REBEL on TACRED and SciERC, and vanilla fine-tuned LLaMA still surpasses REBEL, due to the fact that LLaMA (7B) has a much larger number of model parameters than REBEL (400M). Although LLaMA is smaller than DeepStruct which is based on a pre-trained 10B parameter language model GLM~\cite{du-etal-2022-glm} and is pre-trained on a collection of large-scale corpus, RE$^4$ delivers better results on TACRED than DeepStruct. This is notable because RE$^4$ adopt smaller foundation models for fine-tuning and utilize no external datasets for pre-training compared to DeepStruct. In addition, RE$^4$ with LLaMA achieves comparable results on SemEval and much better results on TACRED compared to ICL-based method GPT-RE. Without fine-tuning, if GPT-3 lacks relevant domain knowledge about specific tasks, then the performance of GPT-RE is limited and greatly affected by the retrieved demonstrations. And this is why GPT-RE delivers exceptional results on SemEval but unattractive results that even worse than LUKE on TACRED. In other words, solely relying on ICL-based methods is unstable. The main results demonstrate the simplicity and effectiveness of RE$^4$ compared to baselines.

\subsection{Discussions}
We conduct extensive experiments to verify the effectiveness of recalling and reasoning module. We also provide detailed analysis about retrieved demonstrations and in-context reasoning abilities. In this section, we only experiment RE$^4$ with LLaMA on SemEval and TACRED for simplicity.

\begin{figure}[t]
\includegraphics[width=0.23\textwidth]{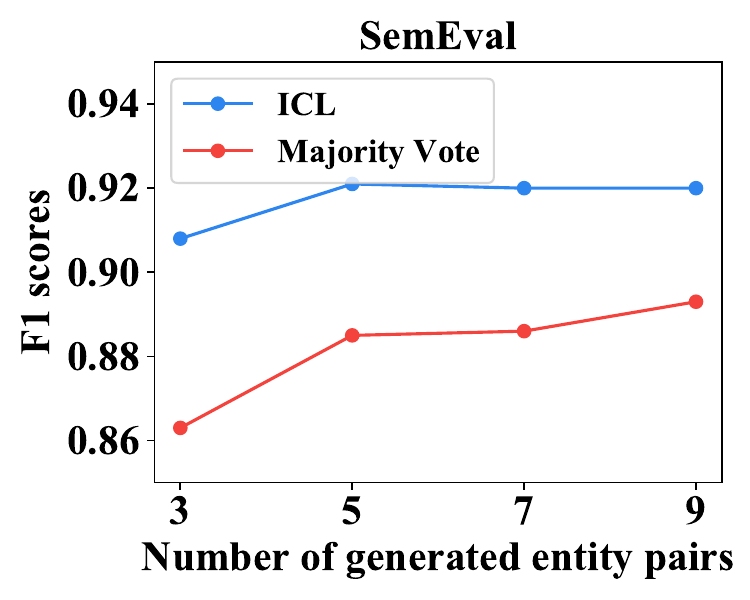} \label{semeval} \hspace{1mm}
\includegraphics[width=0.23\textwidth]{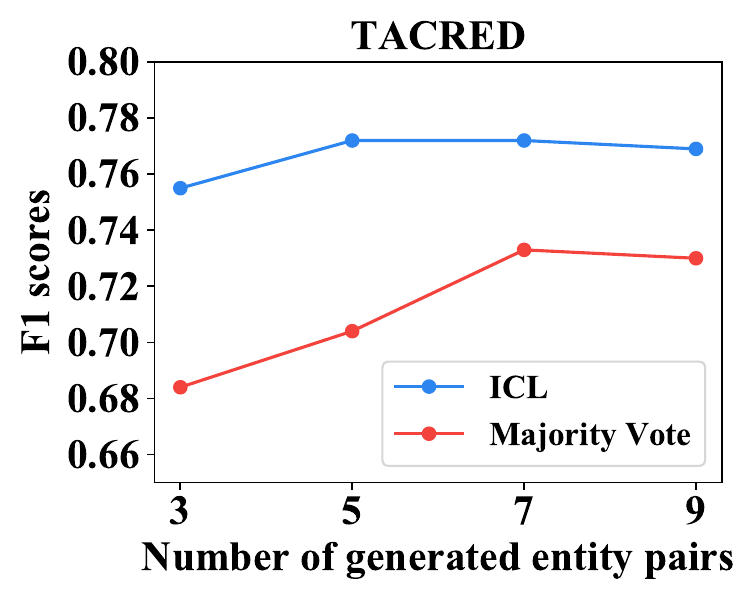} \label{tacred}
\caption{The sensitivity of $k$. \textbf{ICL} denotes we perform in-context reasoning in reasoning module with retrieved demonstrations, while \textbf{Majority Vote} denotes we consider the relation with the maximum number of generated entity pairs as the predicted relation.}
\label{number}
\end{figure}

\paragraph{Number of Generated Entity Pairs.} We vary the number of generated entity pairs $k$ from 3, 5, 7 to 9. We also consider the majority vote results, the overall results are shown in Figure~\ref{number}. On the one hand, we find that without in-context reasoning, the performance are more sensitive to the $k$-selection. With the help of ICL, RE$^4$ could achieve more stable and much better performances, which highlights the importance of reasoning module. On the other hand, obviously, setting $k$ to 5 delivers the best results across two datasets. While increasing $k$ generally boosts the performance of majority vote, the overall performance of RE$^4$ tends to stabilize, even with a slight decrease. This is because in-context reasoning is sensitive to the distractors (i.e., additional retrieved demonstrations that are not relevant to a test example)~\cite{shi2023large}. Therefore, setting larger $k$ not only increases difficulty in generating accurately consistently ontological entities, but brings more distractors that are negative for reasoning module.

\begin{table}[t]
\small
\centering\setlength{\tabcolsep}{3mm}
\begin{tabular}{lcccc}
\toprule
{Dataset} & {\#Test} & {\#Entity Pairs} & {\#Valid} & {\#Ratio} \\
\midrule
{SemEval} & {2,717} & {13,585} & {13,023} & {95.86\%} \\
{TACRED} & {15,509} & {77,545} & {73,482} & {94.76\%} \\
\bottomrule
\end{tabular}
\caption{Validness of generated entity pairs ($k$=5). \#Valid denotes the number of valid generated entity pairs, and \#Ratio denotes the ratio of valid entity pairs in all generated entity pairs.}
\label{valid}
\end{table}

\paragraph{Quality of Generated Entity Pairs.} We consider two aspects of generated entity pairs for quality checking. We first examine the validness of generated entity pairs (i.e., whether it is grounded by retrieval corpora), then we evaluate the relevance of retrieved demonstrations with respect to test examples. We match the generated entity pairs with the entities in training examples, the results are shown in Table~\ref{valid}. It can be seen that under strict exact matching, the proportion of valid generated entity pairs still reaches around 95\%, which indicates that the entity pairs generated by recalling module could serve as the faithful queries for retrieval corpora. However, we are still interested in those invalid entity pairs. Take the SemEval dataset as the example, we analyze the remaining  4.14\% invalid entity pairs, discovering that 70 of 562 entity pairs at least one head or tail entity is grounded by retrieval corpora. The other 492 generated entity pairs all have valid entities, but cannot express a valid relation. Overall, during entity pairs generation, the recalling module reduces the hallucination issue of the LLMs and generates faithful queries.

\begin{table}[t]
\small
\centering\setlength{\tabcolsep}{1mm}
\begin{tabular}{lcccccc}
\toprule
{Dataset} & {5} & {4} & {3} & {2} & {1} & {0} \\
\midrule
{SemEval} & {71.03\%} & {7.45\%} & {6.49\%} & {5.45\%} & {5.21\%} & {4.37\%} \\
{TACRED} & {51.88\%} & {12.75\%} & {5.20\%} & {9.74\%} & {12.26\%} &  {8.17\%} \\
\bottomrule
\end{tabular}
\caption{The number of retrieved examples which share same relation with a test example ($k$ = 5). We consider all the examples in test sets and calculate the proportion.}
\label{same}
\end{table}

\begin{figure}[t]
\includegraphics[width=0.23\textwidth]{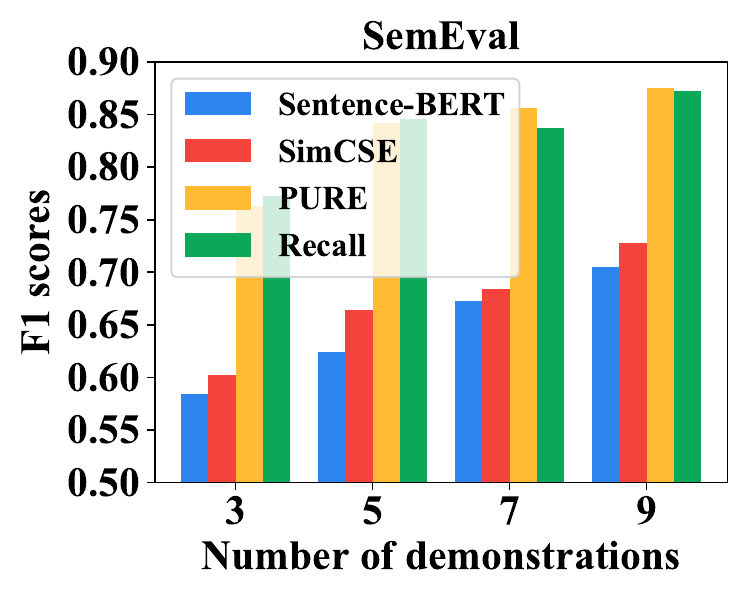} \label{semicl} \hspace{1mm}
\includegraphics[width=0.23\textwidth]{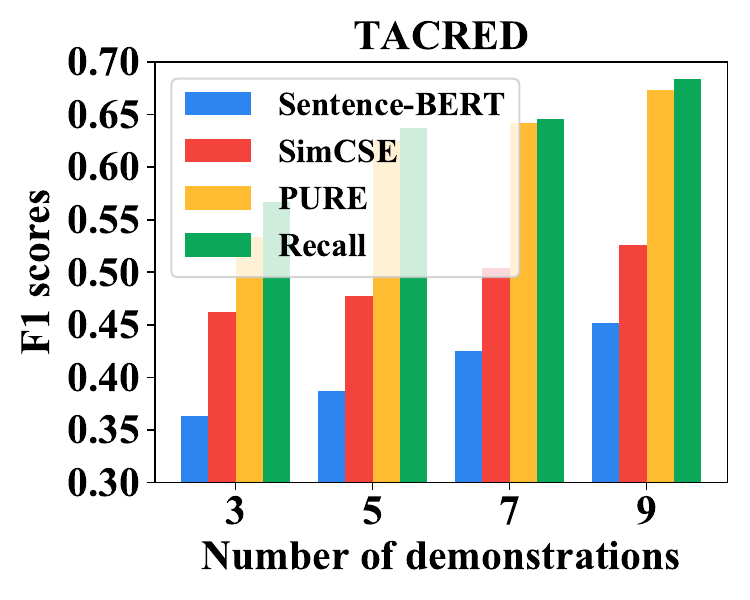} \label{tacicl}
\caption{Comparison on different retrieval models.}
\label{icl}
\end{figure}

To evaluate the relevance of retrieved demonstrations with respect to test examples, we consider the number of examples which share same relation with a test example (i.e. the consistency of ontology). The results are summarized in Table~\ref{same}. For most cases, the entity pairs generated by the recalling module and the demonstrations retrieved by retrieving module are all relevant (i.e., all 5 demonstrations share same relation with a test example). Note that 84.97\% and 69.83\% of retrieved results ensure the majority of golden relation examples participation. And these demonstrations can provide relevant contextual knowledge for ICL. However, one drawback of recalling is that there may be no golden relation in the generated results. Existing similarity-based retrievers such as Sentence-BERT~\cite{reimers2019sentence}, SimCSE~\cite{gao2021simcse} and PURE~\cite{zhong-chen-2021-frustratingly} essentially cannot guarantee that the retrieved demonstrations would always contain the golden relation. The relevance of retrieved demonstrations actually rely on the performance of the retrieval model. Therefore, we evaluate the quality of retrieved demonstrations from recalling module and similarity calculation via ICL on GPT-3~\footnote{For GPT-3, we use ``\texttt{text-davinci-003}''. For Sentence-BERT, we use ``\texttt{all-mpnet-base-v2}''. For SimCSE, we use ``\texttt{sup-simcse-bert-base-uncased}''. For PURE, we use ``\texttt{bert-base-uncased}'' as the backbone.}, as shown in Figure~\ref{icl}. We find that fine-tuned retrievers (PURE and recalling module) substantially achieve much better ICL results than similarity-based models without considering entity and relation semantics. For different number of demonstrations, recalling module delivers comparable results compared to PURE, showing the effectiveness of recalling module and allowing seamless integration with any arbitrary LLMs during inference.

\paragraph{Effectiveness of ICL Tuning.} We remove the recalling instruction task and keep the reasoning instruction task. Not relying on the retrieved demonstrations $\mathcal{Z}^{*}$, we use 5 training examples as demonstrations, sampled uniformly at random, during both training and testing time. Note that we relax the assumption of perfect balance between labels on training examples. We categorize two types of test examples and then compare the performance of RE$^4$ with vanilla fine-tuned LLaMA, the results are shown in Table~\ref{reason}. First, without any tuning process, LLaMA cannot perform ICL in RE. We empirically discover that LLaMA is unable to understand the structural sentences in standard ICL paradigm and recover the relation labels of test examples based on demonstrations. Second, although the distractors can lead LLMs to make inaccurate predictions, the improvement brought by ICL tuning is still obvious. Experimental results suggest that the impact of ICL training is positive when the golden relation is included in retrieved demonstrations compared to vanilla fine-tuning, which also highlights the importance and performance of retrieval models. During instruction tuning, the model learns to learn in-context for deducing the golden relation.

\begin{table}[t]
\small
\centering\setlength{\tabcolsep}{1mm}
\begin{tabular}{lccccccc}
\toprule
\multirow{2}{*}{Training} & \multirow{2}{*}{Inference} & \multicolumn{3}{c}{SemEval} & \multicolumn{3}{c}{TACRED} \\
& & \textbf{I} & \textbf{II} & \textbf{Avg.} & \textbf{I} & \textbf{II} & \textbf{Avg.} \\
\midrule
{w/o tuning}  & {Direct / ICL} & {0.0} & {0.0} & {0.0} & {0.0} & {0.0} & {0.0} \\
{w/ fine-tuning} & {Direct} & {90.7} & \textbf{90.1} & {90.4} & {74.6} & \textbf{75.2} & {74.9} \\
{w/ ICL tuning} & {ICL} & \textbf{91.4} & {89.8} & \textbf{90.6} & \textbf{76.7} & {73.7} & \textbf{75.2} \\
\bottomrule
\end{tabular}
\caption{Comparison between vanilla fine-tuning and ICL tuning. \textbf{I} denotes the model performance on test examples that their demonstrations contain at least one golden relation example, while \textbf{II} denotes the performance when the random selected demonstrations are all distractors. \textbf{Avg.} represents the average score of \textbf{I} and \textbf{II} results.}
\label{reason}
\end{table}

\begin{figure}[t]
\includegraphics[width=0.23\textwidth]{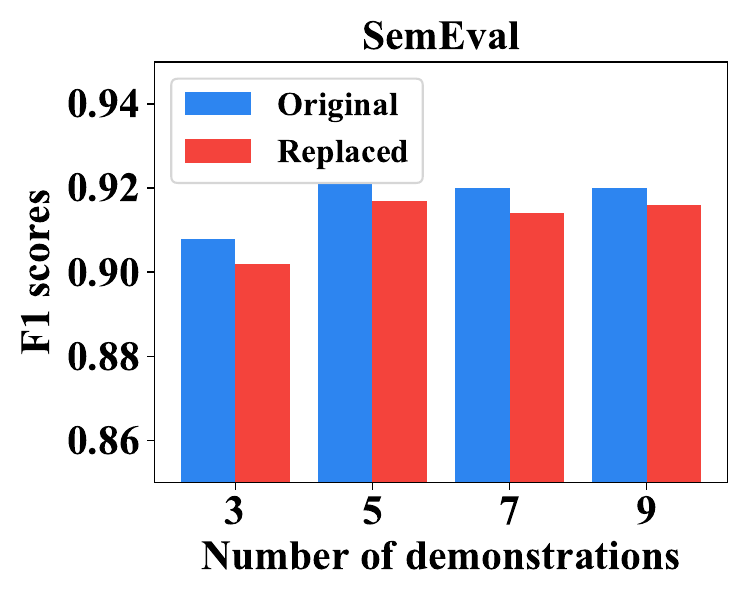} \label{resem} \hspace{1mm}
\includegraphics[width=0.23\textwidth]{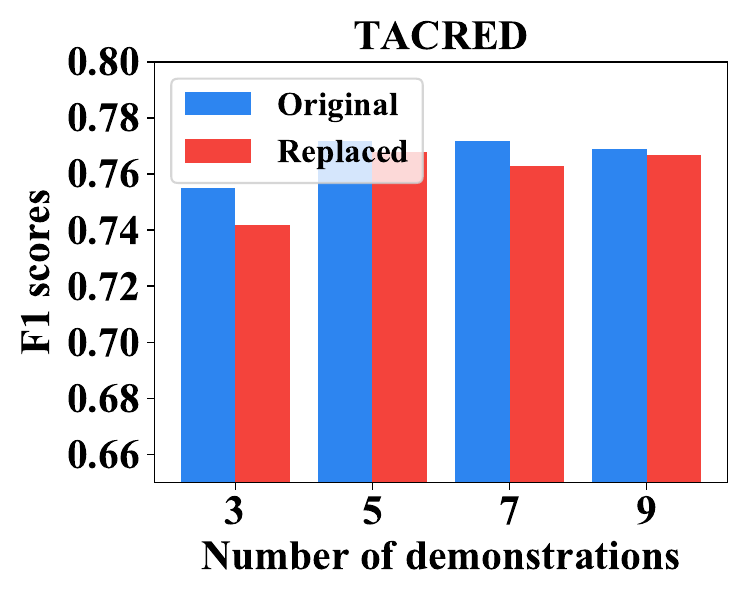} \label{retac}
\caption{Sensitivity of in-context reasoning.}
\label{sen}
\end{figure}

\paragraph{Sensitivity of In-Context Reasoning.} We have validated that relevant contextual knowledge for test examples is beneficial for overall performance. But we might be interested in whether all retrieved demonstrations share similar ontology are equal during ICL. Specifically, for each test example, we replace the corresponding $k$ generated entity pairs with other $k$ random entity pairs but share same relations. The in-context reasoning results are shown in Figure~\ref{sen}. Similar with other LLMs such as GPT-3, the change of demonstrations also have an impact on the final results. Compared to demonstrations obtained via generated entity pairs, the replacement operation consistently brings slightly performance drop across different $k$ and datasets, which indicates that the recalling module actually help to discover similar demonstrations with test examples. However, the performance degradation caused by this randomness is relatively small compared to performance in GPT-3~\cite{zhao2021calibrate}. The ICL tuning process forces the LLMs to deduce the golden relation based on important entities and relations, which indicates that RE$^4$ is able to perform modestly robust reasoning utilizing relevant but imperfect contextual knowledge.

\section{Related Work}
The large language models (LLMs), such as GPT-3~\cite{brown2020language} and ChatGPT~\cite{openai2022}, perform well in various downstream tasks without any training or fine-tuning but only with a few examples as instructions, which is called in-context learning (ICL). However, ICL with LLMs achieves poor performance in relation extraction (RE) where the main obstacles are two-fold: (1) the low relevance regarding entity and relation in the retrieved demonstrations for ICL, and (2) the failure of utilizing LLMs with moderate size for ICL. For the first challenge, existing attempts rely on sentence embedding in retrieval, including the sentence encoders such as Sentence-BERT~\cite{reimers2019sentence} and SimCSE~\cite{gao2021simcse}. Considering entity and relation semantics, GPT-RE~\cite{wan-etal-2023-gpt} fine-tunes PURE~\cite{zhong-chen-2021-frustratingly} to provide more RE-specific and robust representations for retrieval. For the second challenge, although ICL has been further improved by later work~\cite{zhao2021calibrate,holtzman2021surface,min2021noisy} and shows promising results on a variety of tasks, these researches mainly focus on GPT-3. To improve the ICL ability of other LLMs, current methods propose to make LLMs perform better ICL via meta learning~\cite{chen2022meta} and multi-task learning~\cite{min2022metaicl}. Inspired by these, our method can be viewed as the integration framework of retrieving and reasoning by instruction tuning. Guided by ontological knowledge, RE$^4$ generates some possible similar entities and relations, and then performs reliable in-context reasoning based on the retrieved demonstrations.

\section{Conclusion}
In this work, we propose RE$^4$, a novel recall-retrieve-reason RE framework for open-source LLMs, which recalls relevant entity pairs, retrieves good demonstrations and then perform better in-context reasoning for RE. We consider distilling consistently ontological knowledge to guide the demonstration retrieval, and tuning LLMs with ICL objective to perform reliable in-context reasoning. Specially, RE$^4$ allows seamless integration with any arbitrary LLMs during inference. Empirical results show that RE$^4$ achieves new state-of-the-art sentence-level RE performance in four RE benchmarks. We also demonstrate its effectiveness with extensive experiments, and discuss its advantages and limitations, encouraging more effective generative-based RE methods in the future research.

\section*{Acknowledgments}
We thank the reviewers for their insightful comments. This work was supported by National Science Foundation of China (Grant Nos.62376057) and the Start-up Research Fund of Southeast University (RF1028623234). All opinions are of the authors and do not reflect the view of sponsors.

\bibliographystyle{named}
\bibliography{ijcai24}

\end{document}